\pdfoutput=1

\documentclass[11pt]{article}

\usepackage[final]{acl}

\usepackage{times}
\usepackage{latexsym}
\usepackage[T1]{fontenc}
\usepackage[utf8]{inputenc}
\usepackage{microtype}
\usepackage{inconsolata}
\usepackage{graphicx}
\usepackage{paralist, tabularx}
\usepackage{caption}
\usepackage{float}
\usepackage{authblk}
\usepackage{lipsum}
\usepackage{graphicx}
\usepackage{subcaption}
\usepackage{booktabs} 

\newcommand\setalign{4pt}

\title{\includegraphics[scale=0.03]{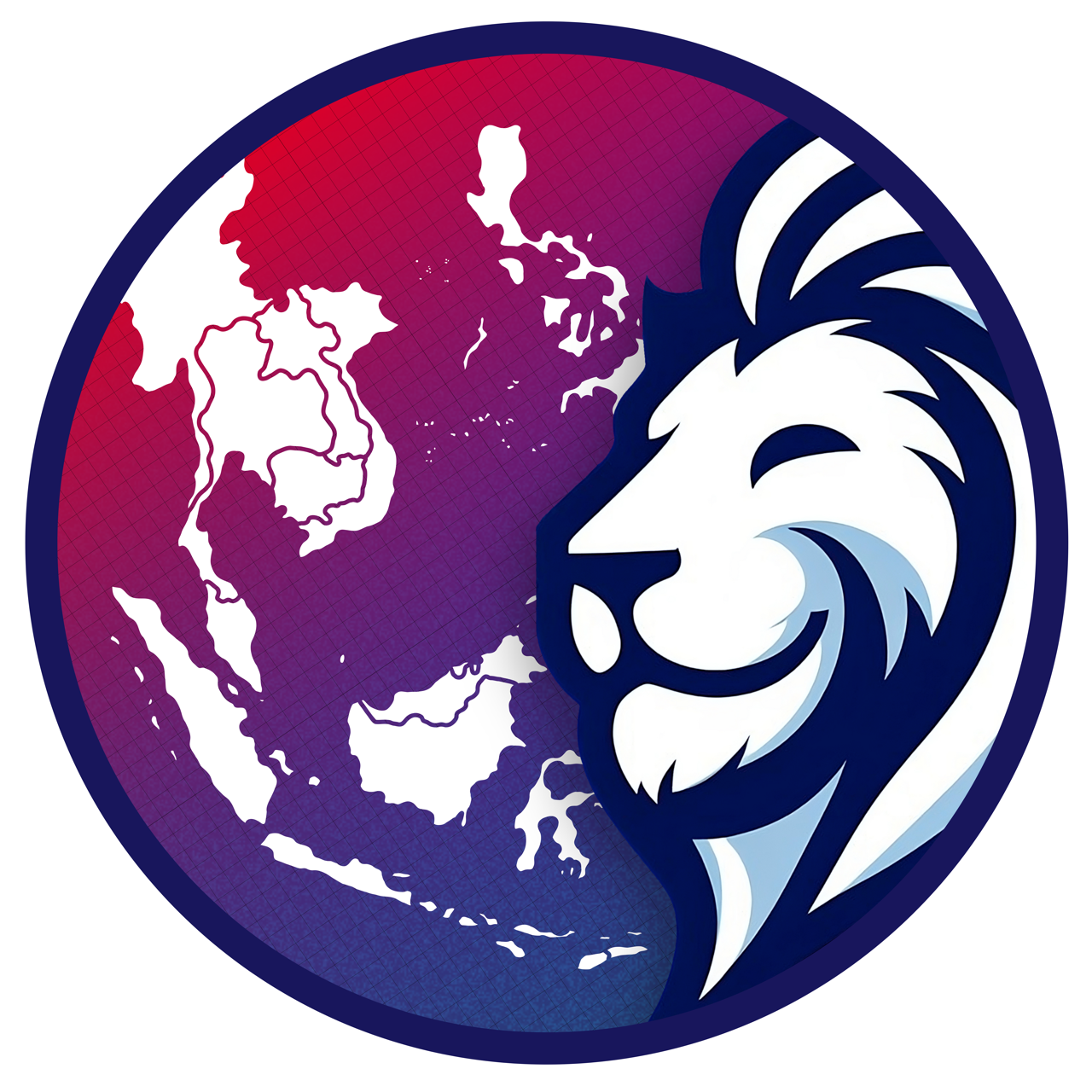} \raisebox{1.5\height}{  SEA-LION: Southeast Asian Languages in One Network} \vspace{-4mm}}

\author{
Raymond Ng\textsuperscript{$\spadesuit$},
Thanh Ngan Nguyen\textsuperscript{$\spadesuit$},
Yuli Huang\textsuperscript{$\spadesuit$},
Ngee Chia Tai\textsuperscript{$\spadesuit$},
Wai Yi Leong\textsuperscript{$\spadesuit$}, \\
Wei Qi Leong\textsuperscript{$\spadesuit$},
Xianbin Yong\textsuperscript{$\spadesuit$},
Jian Gang Ngui\textsuperscript{$\spadesuit$},
Yosephine Susanto\textsuperscript{$\spadesuit$},
Nicholas Cheng\textsuperscript{$\spadesuit$}, \\
Hamsawardhini Rengarajan\textsuperscript{$\spadesuit$},
Peerat Limkonchotiwat\textsuperscript{$\spadesuit$},
Adithya Venkatadri Hulagadri\textsuperscript{$\spadesuit$},
\\
Kok Wai Teng\textsuperscript{$\spadesuit$},
Yeo Yeow Tong\textsuperscript{$\spadesuit$},
Bryan Siow\textsuperscript{$\spadesuit$},
Wei Yi Teo\textsuperscript{$\spadesuit$},
Wayne Lau\textsuperscript{$\spadesuit$},
\\
Choon Meng Tan\textsuperscript{$\spadesuit$},
Brandon Ong\textsuperscript{$\spadesuit$},
Zhi Hao Ong\textsuperscript{$\spadesuit$},
Jann Railey Montalan\textsuperscript{$\spadesuit$},
\\
Adwin Chan\textsuperscript{$\spadesuit$},
Sajeban Antonyrex\textsuperscript{$\spadesuit$},
Ren Lee\textsuperscript{$\spadesuit$},
Esther Choa\textsuperscript{$\spadesuit$},
David Ong Tat-Wee\textsuperscript{$\spadesuit$}, 
\\
Bing Jie Darius Liu\textsuperscript{$\spadesuit$},
William Chandra Tjhi\textsuperscript{$\spadesuit$},
Erik Cambria\textsuperscript{$\diamondsuit$},
Leslie Teo\textsuperscript{$\spadesuit$} \\
  \textsuperscript{$\spadesuit$}AI Singapore, National University of Singapore \\
  \textsuperscript{$\diamondsuit$}Nanyang Technological University\\
  \url{https://sea-lion.ai}
}

\begin{document}
\maketitle

\begin{abstract}
  Recently, Large Language Models (LLMs) have dominated much of the artificial intelligence scene with their ability to process and generate natural languages. 
  However, the majority of LLM research and development remains English-centric, leaving low-resource languages such as those in the Southeast Asian (SEA) region underrepresented. To address this representation gap, we introduce \textbf{Llama-SEA-LION-8B-IT} and \textbf{Gemma-SEA-LION-9B-IT}, two cutting-edge multilingual LLMs designed for SEA languages. The SEA-LION family of LLMs supports 11 SEA languages, namely English, Chinese, Indonesian, Vietnamese, Malay, Thai, Burmese, Lao, Filipino, Tamil, and Khmer. Our work leverages large-scale multilingual continued pre-training with a comprehensive post-training regime involving multiple stages of instruction fine-tuning, alignment, and model merging. Evaluation results on multilingual benchmarks show that our models achieve state-of-the-art performance across LLMs supporting SEA languages. 
  We open-source the models \footnote{\href{https://huggingface.co/collections/aisingapore/sea-lionv3-672589a39cdadd6a5b199581}{SEA-LION Models Collection}} to benefit the wider SEA community.
\end{abstract}


\section{Introduction}

Large language models (LLMs) have significantly transformed the field of natural language processing, achieving remarkable performance in text generation, summarization and sentiment analysis~\citep{DBLP:conf/nips/BrownMRSKDNSSAA20,DBLP:journals/corr/abs-2303-08774,DBLP:journals/corr/abs-2407-21783,DBLP:journals/corr/abs-2408-00118,DBLP:conf/ijcnn/ZhangMC24,DBLP:conf/emnlp/YeoFKSC24}. 
Despite their impressive capabilities, most LLMs remain heavily English-centric~\citep{DBLP:conf/acl/WendlerVM024,DBLP:journals/corr/abs-2408-10811}. 
Unfortunately, this situation has led LLMs in regions with many under-represented languages such as \textbf{S}outh\textbf{e}ast \textbf{A}sia (\textbf{SEA)} to suffer. Languages with lower resources, such as Filipino, Lao, Burmese and Khmer in the SEA region, are not supported by many open-source English-centric LLMs.
%
This underscores the need to bridge the resource and representation gap between English and SEA languages.

Recently, there have been many attempts to create multilingual LLMs in an open-source manner, e.g., BLOOM~\citep{DBLP:journals/corr/abs-2211-05100}, a project aimed at increasing multilingual presence in open-source LLMs by supporting 46 languages.
Popular LLM families such as Llama~\citep{DBLP:journals/corr/abs-2407-21783}, Gemma~\citep{DBLP:journals/corr/abs-2408-00118} and Qwen~\citep{DBLP:journals/corr/abs-2407-10671} have also introduced multilingual LLMs for their latest iteration.
During our evaluations, we found that the performance of these models is acceptable in the general case, i.e., when considering evaluation benchmarks formulated from English datasets. However, we observe that the performance degrades on SEA-specific benchmarks. 
Moreover, researchers have also introduced LLMs such as SeaLLMs~\cite{nguyen-etal-2024-seallms,DBLP:journals/corr/abs-2407-19672} and Sailor~\cite{DBLP:journals/corr/abs-2404-03608} to specifically address the LLM gap in SEA languages. 
However, the performance of these models is less than ideal for languages such as Thai or Tamil\footnote{Tamil is one of the official languages in Singapore. It is also spoken in other areas in the SEA region, such as Malaysia.}~\citep{thaillm-leaderboard, sealion-leaderboard}.
%
%

In this paper, we address the issues by proposing a robust open-source Southeast Asian model with data transparency for reproducibility, namely \textbf{SEA-LION} -- a family of LLMs continued pre-trained (CPT) and fine-tuned on Llama-3.1-8B-Instruct for Llama-SEA-LION-8B-IT and Gemma-2-9B for Gemma-SEA-LION-9B-IT with a focus on SEA languages. 
To tackle the performance problem, we utilize 200 billion English, code, and SEA languages tokens as well as 16.8 million English and SEA languages instruction and answer pairs for CPT and post-training steps, respectively, to achieve a significant improvement in SEA languages.
In order to allow our models to be used by everyone without restrictions, we release our models under the fully open MIT license.
We benchmark our models against the SEA-HELM\citep{susanto2025seahelmsoutheastasianholistic} and Open LLM Leaderboard\footnote{\href{https://huggingface.co/spaces/open-llm-leaderboard/open_llm_leaderboard\#/}{Open LLM Leaderboard}} with other LLMs of similar sizes in Southeast Asia like Sailor 2 \citep{sailor2} and SeaLLMs 3 \citep{DBLP:journals/corr/abs-2407-19672}, where our models achieve state-of-the-art performances.

We summarize the contribution of our paper as follows.
\begin{compactitem}[\hspace{\setalign}•]
    \item {We released two LLMs, \textbf{Llama-SEA-LION-8B-IT} and \textbf{Gemma-SEA-LION-9B-IT}, that are meticulously trained to accurately represent the unique linguistic diversity of SEA languages.}
    \item {We also provide in-depth insights in this paper into our end-to-end training workflow to benefit the community developing multilingual LLMs.}
    \item We present a reproducible dataset development process, covering sourcing and the model training process.
    We release our training artifacts, including the training dataset, training scripts, training checkpoints, and fine-tuned models, including \textbf{Llama-SEA-LION-8B-IT} and \textbf{Gemma-SEA-LION-9B-IT}, to provide strong baselines, promote reproducibility, and enable future research on applications that require SEA-specific knowledge~\footnote{Please visit \url{https://huggingface.co/aisingapore} for all artifacts in this paper, including training data and other versions of SEA-LION}.
\end{compactitem}

$ $

\section{Continued pre-training (CPT)}
\subsection{Pre-training data}
\label{sec:cpt_data}
The CPT data consists of a curated set of English, multilingual, and code corpora from several open source repositories like Dolma~\citep{soldaini-etal-2024-dolma}, FineWeb~\citep{DBLP:journals/corr/abs-2406-17557}, the-stackv2~\citep{DBLP:journals/corr/abs-2402-19173}, SEA-LION-Pile~\citep{sea-lion-pile}, SEA-LION-Pile-v2~\citep{sea-lion-pilev2}, as well as documents from CommonCrawl~\citep{commoncrawl} and from the public domain, such as Wikipedia~\citep{wikimedia}.
For SEA-LION-Pilev2, we filter CommonCrawl WARC data for documents in SEA languages (i.e., Burmese, Simplified Chinese, Indonesian, Khmer, Lao, Malay, Filipino, Tamil, Thai, and Vietnamese) using the pretrained fasttext language classifier~\citep{joulin2017bag}. 

A document is retained if the language code reported in its metadata matches that of one of the aforementioned SEA languages. Additionally, we further clean up the data with Trafilatura~\citep{barbaresi-2021-trafilatura}.
To determine the optimal dataset ratio between SEA languages, code, and English for the CPT process, we conduct a series of small-scale CPT experiments, each with a training budget of 10 billion tokens and varying proportions of English, code, and SEA language data.
We settled on an optimal data mix ratio of 55\% SEA languages, 25\% English, and 20\% code tokens for a budget of 200 billion tokens. 
For a detailed breakdown of the token count by languages, please refer to Table~\ref{tab:cpt-data}.

\subsection{CPT process}
\label{sec:cpt_proc}

\noindent
\textbf{Model selection}.
For the models to CPT from, we choose Llama-3.1-8B-Instruct~\citep{DBLP:journals/corr/abs-2407-21783} and Gemma-2-9B~\citep{DBLP:journals/corr/abs-2408-00118}.


\noindent
\textbf{Training setup}.
Following previous works~\cite{DBLP:journals/corr/abs-2404-03608}, we use BPE-Dropout~\citep{DBLP:conf/acl/ProvilkovEV20} to increase the performance and robustness of the training. 
We use a Warmup-Stable-Decay (WSD)~\citep{DBLP:journals/corr/abs-2404-06395} scheduler with warm-up and cooldown phases each representing 10\% of the entire training budget. 
We use the AdamW~\citep{DBLP:conf/iclr/LoshchilovH19} optimizer with the maximum learning rate (LR) set to ${1e^{-5}}$ and the final LR after cooldown is ${1e^{-7}}$. Following~\citet{DBLP:conf/iclr/WortsmanLXEAACG24}, we set epsilon to ${1e^{-15}}$.
We use Composer~\citep{mosaicml2022composer} and LLM Foundry~\citep{mosaicml2022llmfoundry} for distributed training using Fully Sharded Data Parallel~\citep{DBLP:journals/pvldb/ZhaoGVLHXWSOSDB23} on a cluster of eight nodes of the p5.48xlarge instance from Amazon Web Services (AWS). 
The total training duration was approximately 6 days and 10 days for the Llama 3.1 and Gemma 2 models, respectively.
In this paper, we refer to the post-CPT models as \emph{Llama-SEA-LION-8B} and \emph{Gemma-SEA-LION-9B} for the Llama 3.1 and Gemma 2 continued pre-trained models, respectively.

\section{Post-training}
\subsection{Post-training data}
\label{sec:sft_data}
The post-training data consists of 3 subsets of data for Stage 1 IFT, Stage 2 IFT, and the Preference dataset for alignment, respectively.
We describe the training data information of each step as follows.

\noindent
\textbf{Stage 1 IFT}. 
In this step, we employ Infinity-Instruct [Foundation and Chat]~\citep{InfinityInstruct2024} and OpenMath-Instruct 2~\citep{DBLP:journals/corr/abs-2410-01560} to improve the mathematical, reasoning, and coding skills of the instruction model.
The full details of the training data are shown in Appendix~\ref{tab:stage-1-ift-data}.

\noindent
\textbf{Stage 2 IFT}. Then, in this step, we use generalized large-scale instructions on the previous instruction model.
In particular, we employ 22 existing datasets (written in English, Thai, and Vietnamese) and formulate new 22 synthetic datasets using various models and techniques to create SEA instruction datasets (see Appendix~\ref{sec:stage-2-data} for the full data generation details). 
As shown in Appendix~\ref{tab:stage-2-ift-data}, we use a total of 7,298,828 instruction samples that cover 11 languages.

\noindent
\textbf{Helpfulness and preference alignment}. We also conduct an alignment learning on top of the instruction model using a feedback dataset called UltraFeedBack~\cite{DBLP:conf/icml/CuiY0YH0NXXL0024}.
In addition, we also synthesized the SEA version of the UltraFeedBack using NemoTron-70b with Gemma2 as a reward model, see Appendix~\ref{sec:helpfulness-and-preference-alignment-data} for the full details.


\begin{figure}[h]
 \includegraphics[width=\columnwidth]{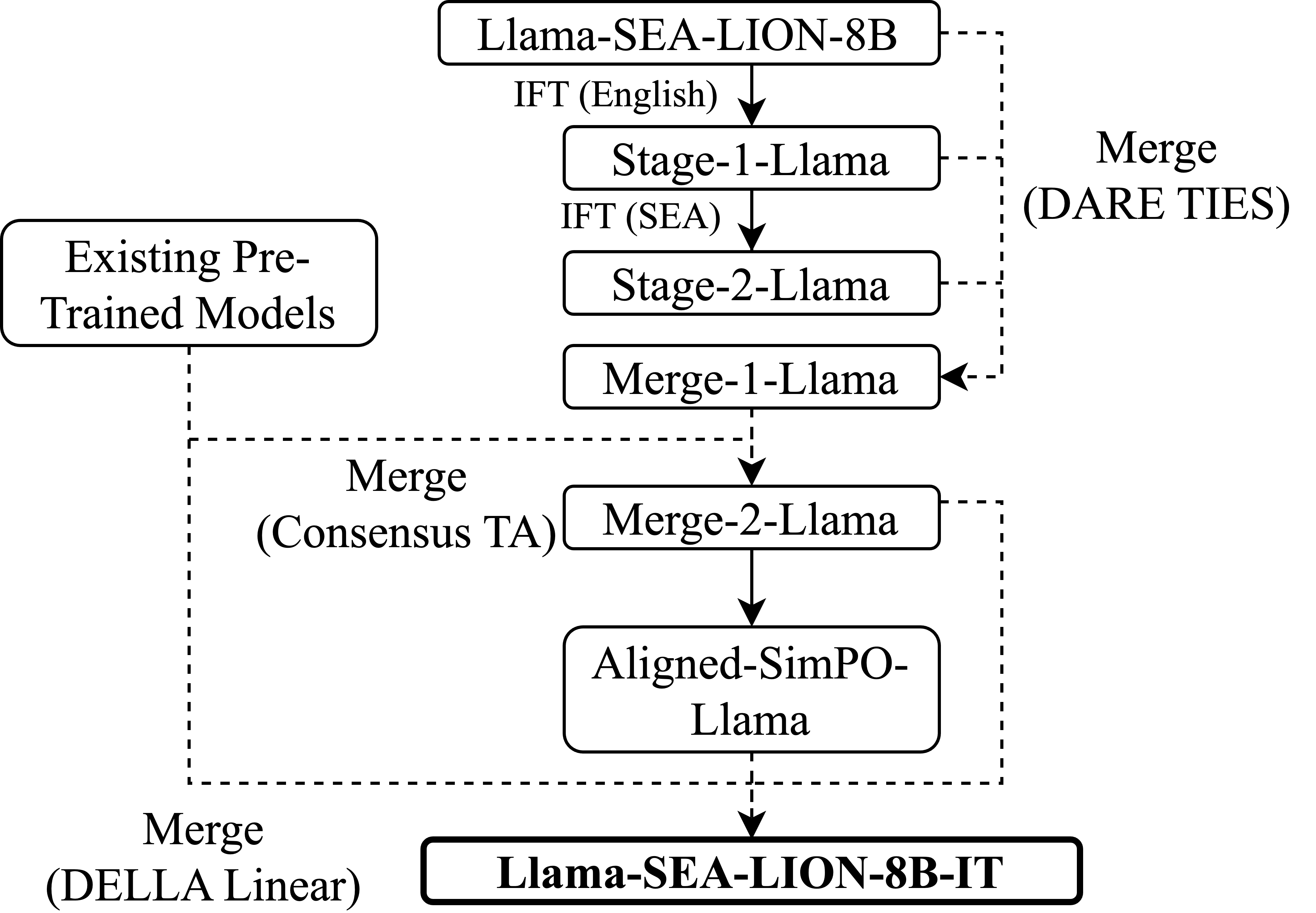}
 \centering
 \caption{Training process of \textbf{Llama-SEA-LION-8B-IT} (Section~\ref{sec:sft_llama_proc}). The post-training process consists of 2 stages of instruction fine-tuning, an alignment stage and multiple merge stages. Dotted lines denote a merge stage and solid lines denote an alignment stage.}
 \label{fig:sealionv3-llama-training}
 \vspace{-3mm}
\end{figure}

\subsection{Post-training process}
\label{sec:sft_proc}
%
We use LLaMaFactory~\citep{zheng2024llamafactory} with DeepSpeed~\citep{DBLP:conf/kdd/RasleyRRH20} for all Instruction Fine Tuning (IFT) and alignment steps. 
All IFT stages are performed using full model fine-tuning, where the models are from the previous step (Section~\ref{sec:cpt_proc}) and existing models. 
We use MergeKit~\citep{DBLP:conf/emnlp/GoddardSEMKBMS24} with a value of 1 for weight and density parameters for all merge steps. 
Models selected for merging are selected empirically, based on the openness of model licenses, the suitability for merging and performance.

\subsubsection{Llama-SEA-LION-8B-IT}
\label{sec:sft_llama_proc}

\noindent
\textbf{Stage 1 IFT}
\label{sec:ift_stage1}
As shown in Figure~\ref{fig:sealionv3-llama-training}, we started off the post-training phase with IFT of \emph{Llama-SEA-LION-8B} with the Infinity Instruct (Foundation)~\citep{InfinityInstruct2024} and OpenMathInstruct2~\cite{DBLP:journals/corr/abs-2410-01560} datasets. Both datasets contain approximately 9.5 million instruction pairs, primarily in English and centered around reasoning, math, and code. We refer to the model at this stage as \emph{Stage-1-Llama}.

\noindent
\textbf{Stage 2 IFT}
\label{sec:ift_stage2}
We performed a second round of IFT using the SEA-Instruct dataset, which consists of approximately 7.3 million instruction pairs, of which 5 million instruction pairs are generated using the Gemma-2-27B-Instruct~\citep{DBLP:journals/corr/abs-2408-00118} model and the Qwen2.5-32B-Instruct model~\cite{DBLP:journals/corr/abs-2407-10671} in SEA languages. The remaining are English language instruction pairs from the Infinity-Instruct (Chat)~\cite{InfinityInstruct2024} dataset. We refer to the model at this stage as \emph{Stage-2-Llama}.

\noindent
\textbf{First merge}
After finishing the IFT stages, we performed the first of a series of merges by merging \emph{Stage-1-Llama} and \emph{Stage-2-Llama} into the \emph{Llama-SEA-LION-8B} using the DARE TIES~\citep{DBLP:conf/icml/Yu0Y0L24,DBLP:conf/iclr/IlharcoRWSHF23} method. We refer to the model at this stage as \emph{Merge-1-Llama}.

\noindent
\textbf{Second merge}
In order to mitigate catastrophic forgetting due to the fine-tuning process~\citep{alexandrov-etal-2024-mitigating}, we performed the second round of merging by merging top-performing instruction-tuned models that share the Llama 3.1 lineage. We merge the original Llama-3.1-8B-Instruct, Llama3-8B-SEA-LION-v2.1-Instruct~\citep{sealionv2}, and SuperNova-Lite~\citep{supernovalite} into \emph{Merge-1-Llama} using the Consensus TA~\citep{DBLP:conf/icml/WangDOFF24,DBLP:conf/iclr/IlharcoRWSHF23} merge method. We refer to the model at this stage as \emph{Merge-2-Llama}.

\noindent
\textbf{Helpfulness and preference alignment} We performed one round of alignment on \emph{Merge-2-Llama} using SimPO~\citep{DBLP:journals/corr/abs-2405-14734} with the SEA-Preference dataset. We refer to the model at this stage as \emph{Aligned-SimPO-Llama}.

\noindent
\textbf{Final merge}
Lastly, we perform a merge using the DELLA-Linear merge. With the original Llama-3.1-8B-Instruct model as the base for merging, we merge in \emph{Merge-2-Llama} and \emph{Aligned-SimPO-Llama} to produce the final model, \textit{\textbf{Llama-SEA-LION-v3-9B-IT}}.


$ $

\subsubsection{Gemma-SEA-LION-9B-IT}
\label{sec:sft_gemma_proc}

\begin{figure}[h]
 \includegraphics[width=\columnwidth]{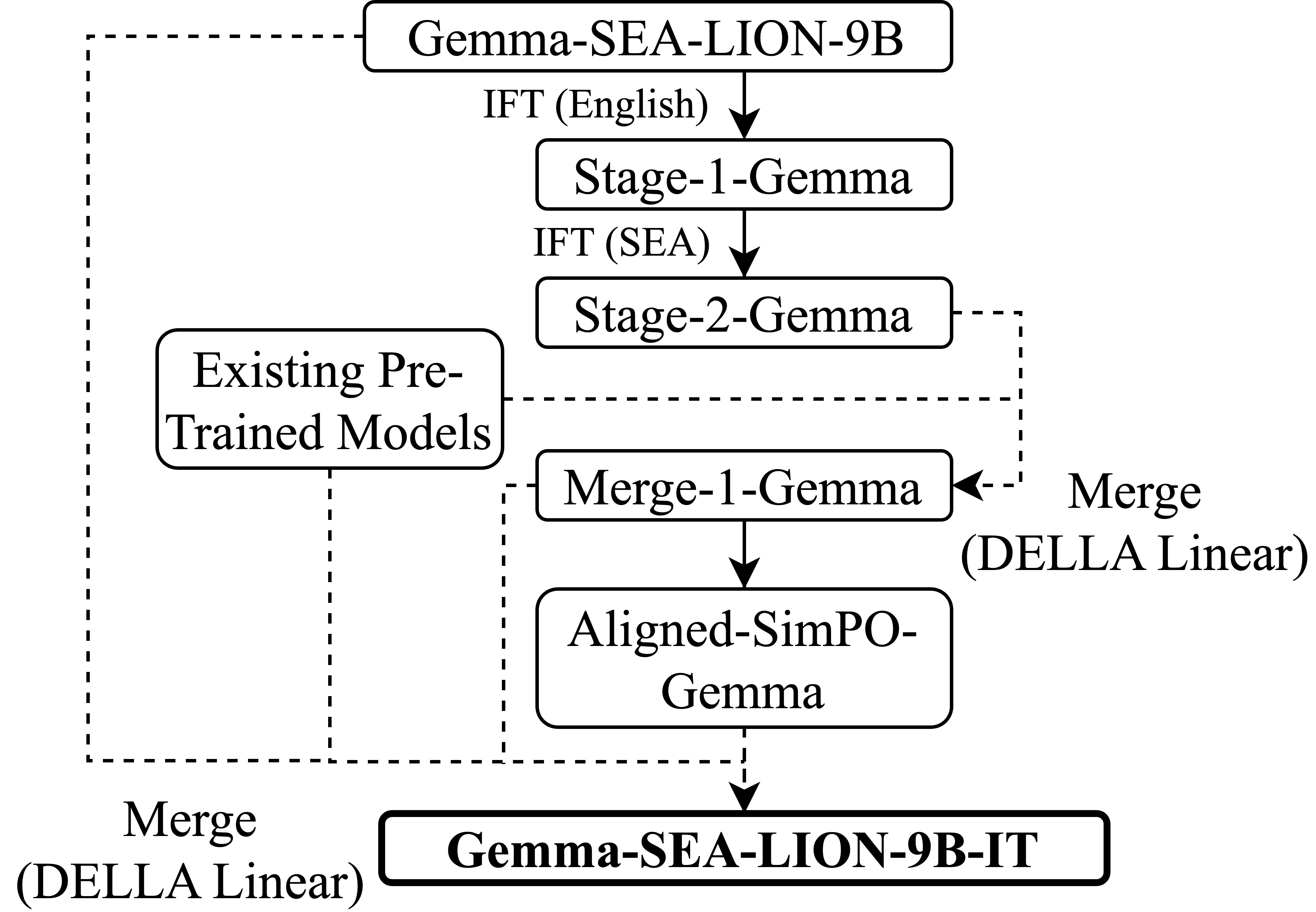}
 \centering
 \caption{Training process of Gemma-SEA-LION-9B-IT (Section~\ref{sec:sft_gemma_proc}). The post-training process comprises two stages of instruction fine-tuning, an alignment stage, and multiple merge stages. Dotted lines denote a merge stage and solid lines denote an alignment stage.}
 \label{fig:sealionv3-gemma-training}
 \vspace{-3mm}
\end{figure}

\noindent
\textbf{Stage 1 and Stage 2 IFT}
\label{sec:gemma_ift_stage1n2}
Similar to the \textit{Llama-SEA-LION-8B-IT}, we started off the post-training phase with both stages of IFT using the same datasets on the Gemma-2-9B model~\citep{DBLP:journals/corr/abs-2408-00118}. We refer to both models at stage 1 and stage 2 as \emph{Stage-1-Gemma} and \emph{Stage-2-Gemma}, respectively.

\noindent
\textbf{First merge}
We merge the Gemma-2-9B-IT~\citep{DBLP:journals/corr/abs-2408-00118} and \emph{Stage-2-Gemma} into Gemma-2-9B using the DELLA Linear method. We refer to the model at this stage as the \emph{Merge-1-Gemma}. \\
\noindent
\textbf{Helpfulness and preference alignment}
Using the \emph{Merge-1-Gemma} as the base model, we performed one round of alignment using SimPO with the SEA-Preference dataset. We refer to the model at this stage as the \emph{Aligned-SimPO-Gemma}. \\
\noindent
\textbf{Final merge}
Finally, using the Gemma-2-9B model as the base model, we merged \emph{Merge-1-Gemma}, FuseChat Gemma-2-9B-Instruct~\citep{DBLP:journals/corr/abs-2412-03187}, \emph{Gemma-SEA-LION-9B}, and \emph{Aligned-SimPO-Gemma} into it to produce the final model \textit{\textbf{Gemma-SEA-LION-9B-IT}}. 

\subsection{Discussion}
This post-training workflow emphasizes the careful balance between general capabilities, SEA-specific linguistic fluency, and natural conversational abilities. Each step in the workflow is designed to progressively refine the model, ensuring it meets the diverse needs of users in the Southeast Asian region.

The entire post-training process for \textit{\textbf{Gemma-SEA-LION-9B-IT}} and \textit{\textbf{Llama-SEA-LION-8B-IT}} took approximately 1350 and 1024 GPU hours, respectively, on eight H100 GPUs. 
To make the training efficient, all post-training steps utilize Liger Kernel~\citep{hsu2024ligerkernelefficienttriton} for substantial memory savings of approximately 60\%.

\section{Experimental Setup}
%

\subsection{Competitive methods}

For the evaluation, we compared our models against well-known LLMs for multilingual and SEA languages, such as \textit{SeaLLMsv3}~\cite{DBLP:journals/corr/abs-2407-19672}, \textit{Sailorv2}~\cite{sailor2}, \textit{Qwen 2.5}~\cite{DBLP:journals/corr/abs-2407-10671}, \textit{Gemma 2}~\cite{DBLP:journals/corr/abs-2408-00118} and \textit{Llama 3.1}~\cite{DBLP:journals/corr/abs-2407-21783}, where the parameters of those models are less than 10 billion parameters, similar to our models. 
%
%

\begin{table*}[ht]
\captionsetup{justification=centering}
\resizebox{\textwidth}{!}{
\begin{tabular}{lccccccccc}
\multicolumn{10}{c}{\textbf{SEA-HELM}} \\
\hline
&         & \multicolumn{4}{c}{\textbf{NLU, NLG, NLR, NLI}}    & \multicolumn{4}{c}{\textbf{Instruction Following}}  \\
\hline
\textbf{Models}   & \textbf{Average} & \textbf{ID}    & \textbf{VI}    & \textbf{TH}    & \textbf{TA}    & \textbf{ID}    & \textbf{VI}    & \textbf{TH}    \\
\hline
Meta-Llama-3.1-8B & 35.37            & 42.33          & 40.67          & 35.13          & 38.88          & 16.19          & 19.05          & 9.00           \\
SeaLLMs-v3-7B     & 37.04            & 44.79          & 48.29          & 43.53          & 27.45          & 26.67          & 35.24          & 26.00          \\
Gemma-2-9B        & 41.48            & 47.65          & 43.28          & 42.00          & 53.26          & 4.76           & 3.81           & 10.00          \\
Qwen2.5-7B        & 41.98            & 51.63          & \textbf{52.17} & 46.55          & 36.60          & \textbf{31.43} & \textbf{36.19} & 30.00          \\
Sailor2-8B        & 42.62            & 53.23          & 47.33          & 46.64          & 45.04          & 30.48          & 30.48          & \textbf{35.00} \\
\hline
Llama-SEA-LION-8B         & 41.42            & 44.98          & 46.25          & 42.79          & 43.03          & 25.71          & 32.38          & 23.00          \\
Gemma-SEA-LION-9B         & \textbf{48.67}   & \textbf{57.16} & 49.39          & \textbf{47.16} & \textbf{60.56} & 25.71          & 20.00          & 27.00          \\
\hline
\end{tabular}
}
\caption{SEA-HELM multilingual benchmark on NLU, NLG, NLR, NLI and instruction following on \underline{base} and \underline{continued pre-trained models} of similar sizes.}
\label{tab:cpt-sea-helm-table}
\end{table*}

\begin{table*}[ht]
\captionsetup{justification=centering}
\resizebox{\textwidth}{!}{
\begin{tabular}{lccccccc}
\multicolumn{8}{c}{\textbf{Open LLM Leaderboard}}                                                    \\
\hline
\textbf{Models}  & \textbf{Average} & \textbf{MMLU-PRO} & \textbf{BBH}  & \textbf{GPQA} & \textbf{MATH Lvl 5} & \textbf{IFEval (EN)} & \textbf{MUSR} \\
\hline
Meta-Llama-3.1-8B & 13.9       & 24.95       & 25.29     & 6.32      & 5.14        & 12.7         & 8.98      \\
Sailor2-8B    & 17.71      & 25.74       & 27.62     & 4.87      & 7.02        & 21.95        & \textbf{19.03} \\
Gemma-2-9B    & 21.15      & 34.48       & 34.1      & \textbf{10.51} & 13.14        & 20.4         & 14.3      \\
SeaLLMs-v3-7B   & 24.00      & 35.71       & 34.57     & 9.28      & 18.81        & 32.94        & 12.68     \\
Qwen2.5-7B    & \textbf{24.99}  & \textbf{37.39}  & 35.81     & 9.96      & \textbf{18.88}        & \textbf{33.74}    & 14.14     \\
\hline
Llama-SEA-LION-8B     & 16.61      & 27.6       & 26.04     & 7.49      & 9.89        & 16.56        & 12.07     \\
Gemma-SEA-LION-9B     & 22.41      & 32.78       & \textbf{37.24} & 10.29     & 9.89        & 30.12        & 14.11     \\
\hline
\end{tabular}
}
\caption{Open LLM Leaderboard benchmarks across different \underline{continued pre-trained models} of similar sizes.}
\label{tab:base-lm-eval-table}
\end{table*}

\subsection{Evaluation Benchmarks}
To evaluate the robustness of our proposed models, we compare our models to competitors in three benchmarks.

\noindent
\textbf{SEA Benchmarks}. We evaluated the multilingual performance of each LLM using the \emph{SEA-HELM Leaderboard}~\citep{DBLP:journals/corr/abs-2309-06085, susanto2025seahelmsoutheastasianholistic}~\footnote{Please visit \url{https://leaderboard.sea-lion.ai/} for live score update of SEA-LION.}. 
%
%
We selected SEA-HELM because the design choice of this benchmark reflects the performance of SEA culture and knowledge the most compared with other existing benchmarks~\cite{seaexam2024,DBLP:conf/emnlp/LoveniaMAMSAFMI24,DBLP:conf/naacl/WangLHJDAC24}.
We also evaluate on a wide-range SEA coverage language benchmark called \emph{SEACrowd}~\cite{DBLP:conf/emnlp/LoveniaMAMSAFMI24}. 
This benchmark consists of all SEA languages for natural language understanding and generation datasets.
However, due to maintenance reasons, we cannot reproduce the NLG benchmark of SEACrowd.
Therefore, we experiment only with the NLU benchmark (zero-shot), which has 131 data subsets, 7 tasks, and 31 SEA indigenous languages.


\noindent
\textbf{English performance}. We also evaluated the English performance of the models using the Open LLM Leaderboard~\citep{openllmleaderboard}. 
This is because English is also widely used in SEA countries.
Therefore, we need to evaluate the understanding and knowledge of LLMs in the English benchmark as well.
The leaderboard consists of six benchmarks, IFEval~\cite{DBLP:journals/corr/abs-2311-07911}, Big Bench Hard~\citep{DBLP:conf/acl/SuzgunSSGTCCLCZ23}, MATH~\citep{DBLP:conf/nips/HendrycksBKABTS21}, GPQA~\citep{DBLP:journals/corr/abs-2311-12022}, MuSR~\citep{DBLP:conf/iclr/SpragueYBCD24} and MMLU-PRO~\citep{DBLP:journals/corr/abs-2406-01574}. 
Moreover, we also evaluate the CPT models on SEA-HELM and the Open LLM Leaderboard since these benchmarks support the CPT evaluation.

\section{Experimental Results}

\label{sec:cpt_benchmarks}
%

\begin{table*}[ht]
\centering
\captionsetup{justification=centering}
\resizebox{\textwidth}{!}{
\begin{tabular}{lccccccccccc}
\multicolumn{12}{c}{\textbf{SEA-HELM}}                                                                                                                                                                                                                                              \\ \hline
\textbf{}               & \multicolumn{1}{l|}{\textbf{}}        & \multicolumn{4}{c|}{\textbf{NLU, NLG, NLR, NLI}}                                       & \multicolumn{3}{c|}{\textbf{Instruction Following}}                   & \multicolumn{3}{c}{\textbf{MTBench}}             \\ \hline
\textbf{Models}         & \multicolumn{1}{c|}{\textbf{Average}} & \textbf{ID}    & \textbf{VI}    & \textbf{TH}    & \multicolumn{1}{c|}{\textbf{TA}}    & \textbf{ID}    & \textbf{VI}    & \multicolumn{1}{c|}{\textbf{TH}}    & \textbf{ID}    & \textbf{VI}    & \textbf{TH}    \\ \hline
SeaLLMs-v3-7B-Chat      & \multicolumn{1}{c|}{39.19}            & 42.72          & 48.50          & 42.59          & \multicolumn{1}{c|}{12.06}          & 57.14          & 53.33          & \multicolumn{1}{c|}{47.00}          & 59.81          & 65.24          & 56.59          \\
Llama-3.1-8B-Instruct   & \multicolumn{1}{c|}{41.48}            & 51.50          & 51.31          & 45.32          & \multicolumn{1}{c|}{15.40}          & 77.14          & 75.24          & \multicolumn{1}{c|}{63.00}          & 56.38          & 57.59          & 54.34          \\
Sailor2-8B-Chat         & \multicolumn{1}{c|}{43.13}            & 48.98          & 48.01          & 45.44          & \multicolumn{1}{c|}{28.29}          & 49.52          & 45.71          & \multicolumn{1}{c|}{40.00}          & \textbf{69.76} & 66.97          & \textbf{73.94} \\
Qwen2.5-7B-Instruct     & \multicolumn{1}{c|}{44.58}            & 60.28          & 53.46          & 53.43          & \multicolumn{1}{c|}{21.03}          & 81.90          & 69.52          & \multicolumn{1}{c|}{66.00}          & 65.66          & 66.80          & 68.71          \\
Gemma-2-9B-IT           & \multicolumn{1}{c|}{55.33}            & 64.04          & 59.86          & 57.22          & \multicolumn{1}{c|}{52.28}          & 88.57          & 78.10          & \multicolumn{1}{c|}{71.00}          & 68.78          & 68.37          & 73.51          \\ \hline
Stage-1-Llama           & \multicolumn{1}{c|}{50.76}            & 51.84          & 51.83          & 46.23          & \multicolumn{1}{c|}{27.53}          & 69.52          & 73.33          & \multicolumn{1}{c|}{59.00}          & 42.74          & 46.41          & 46.46          \\
Stage-2-Llama           & \multicolumn{1}{c|}{59.49}            & 53.87          & 55.18          & 50.92          & \multicolumn{1}{c|}{44.80}          & 77.14          & 76.19          & \multicolumn{1}{c|}{67.00}          & 50.90          & 53.72          & 46.97          \\
Merge-1-Llama           & \multicolumn{1}{c|}{59.36}            & 56.73          & 56.82          & 51.71          & \multicolumn{1}{c|}{46.63}          & 81.90          & 82.86          & \multicolumn{1}{c|}{67.00}          & 57.04          & 54.01          & 50.28          \\
Merge-2-Llama           & \multicolumn{1}{c|}{58.01}            & 59.19          & 52.63          & 51.89          & \multicolumn{1}{c|}{35.40}          & 87.62          & 80.95          & \multicolumn{1}{c|}{78.00}          & 56.38          & 59.32          & 58.86          \\
Aligned-SimPO-Llama     & \multicolumn{1}{c|}{51.30}            & 54.86          & 51.69          & 46.77          & \multicolumn{1}{c|}{26.40}          & 82.86          & 80.00          & \multicolumn{1}{c|}{68.00}          & 68.20          & 64.68          & 64.92          \\ \hline
Llama-SEA-LION-8B-IT & \multicolumn{1}{c|}{61.84}            & 60.50          & 61.48          & 55.92          & \multicolumn{1}{c|}{43.61}          & 84.76          & 85.71          & \multicolumn{1}{c|}{76.00}          & 62.65          & 68.32          & 65.13          \\ \hline
Stage-1-Gemma           & \multicolumn{1}{c|}{56.56}            & 55.06          & 54.51          & 51.96          & \multicolumn{1}{c|}{42.74}          & 66.67          & 74.29          & \multicolumn{1}{c|}{61.00}          & 47.35          & 47.26          & 55.05          \\
Stage-2-Gemma           & \multicolumn{1}{c|}{66.66}            & 64.10          & 61.76          & 56.90          & \multicolumn{1}{c|}{57.85}          & 89.52          & 82.86          & \multicolumn{1}{c|}{76.00}          & 60.54          & 58.93          & 58.76          \\
Merge-1-Gemma           & \multicolumn{1}{c|}{69.26}            & 66.25          & 64.95          & \textbf{59.74} & \multicolumn{1}{c|}{\textbf{60.41}} & 89.52          & \textbf{91.43} & \multicolumn{1}{c|}{\textbf{82.00}} & 66.45          & 64.47          & 65.00          \\
Aligned-SimPO-Gemma     & \multicolumn{1}{c|}{\textbf{69.37}}   & 65.69          & \textbf{65.47} & 59.51          & \multicolumn{1}{c|}{57.38}          & 86.67          & 88.57          & \multicolumn{1}{c|}{78.00}          & 68.89          & \textbf{73.67} & 73.51          \\ \hline
Gemma-SEA-LION-9B-IT & \multicolumn{1}{c|}{69.35}            & \textbf{66.26} & 64.93          & 59.23          & \multicolumn{1}{c|}{58.82}          & \textbf{94.29} & 88.57          & \multicolumn{1}{c|}{78.00}          & 65.85          & 73.27          & 69.07     \\ \hline  
\end{tabular}
}\caption{SEA-HELM multilingual benchmark on NLU, NLG, NLR, NLI, instruction following and multi-turn chat on instruct models of similar sizes.}
\label{tab:seahelm-posttrain-table}
\end{table*}

\begin{figure*}[t]
    \centering
    \begin{minipage}[b]{0.75\linewidth}
        \centering
        \begin{subfigure}[b]{\linewidth}
            \centering
            \includegraphics[width=\linewidth,trim={0, 0, 0, 0}, clip]{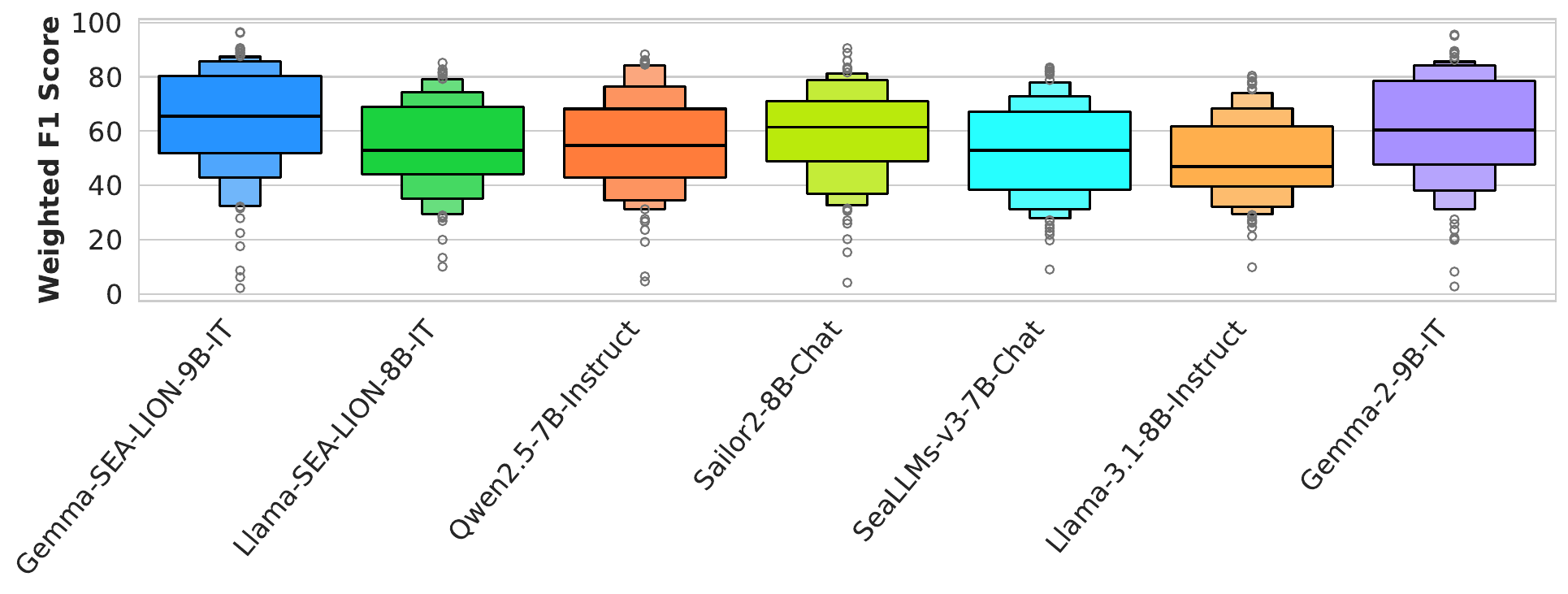}
            
        \end{subfigure}
        \caption{Zero-shot model performance across NLU tasks in SEA languages.}
        \label{fig:sea-nlu-overall-results}
    \end{minipage}
    \hfill
    \begin{minipage}[b]{0.23\linewidth}
    \centering
        \resizebox{1\linewidth}{!}{
        \begin{tabular}{l p{1.2cm}}
          \toprule
          \textbf{Model} & \textbf{NLU Score} \\
          \toprule
            SeaLLMs-v3-7B-chat & 52.68 \\
            Llama-3.1-8B-Instruct & 49.94 \\
            Sailor2-8B-Chat & 60.21 \\
            Qwen2.5-7B-Instruct & 54.51 \\
            Gemma-2-9B-IT & 60.21 \\
            Llama-SEA-LION-8B-IT & 55.10 \\
            Gemma-SEA-LION-9B-IT & \textbf{{64.13}} \\
          \bottomrule
        \end{tabular}
        }
        \captionof{table}{The average NLU performance across 131 data subsets and 31 indigenous languages.}
        \label{tab:lang-equity}
        \vspace{-6pt}
    \end{minipage}
\end{figure*}

\begin{table*}[ht]
\captionsetup{justification=centering}
\resizebox{\textwidth}{!}{
\begin{tabular}{lccccccc}
\multicolumn{8}{c}{\textbf{Open LLM Leaderboard}}                                                                 \\
\hline
\textbf{Models}        & \multicolumn{1}{l}{\textbf{Average}} & \multicolumn{1}{l}{\textbf{MMLU-PRO}} & \multicolumn{1}{l}{\textbf{BBH}} & \multicolumn{1}{l}{\textbf{GPQA}} & \multicolumn{1}{l}{\textbf{MATH Lvl 5}} & \multicolumn{1}{l}{\textbf{IFEval (EN)}} & \multicolumn{1}{l}{\textbf{MUSR}} \\
\hline
Sailor2-8B-Chat        & 16.37                & 27.93                 & 27.15              & 3.47               & 0.00                  & 37.49                  & 2.19               \\
SeaLLMs-v3-7B-Chat      & 22.49                & 33.93                 & 24.37              & 7.27               & 15.86                  & 44.10                  & 9.38               \\
Llama-3.1-8B-Instruct     & 27.88                & 29.36                 & 26.10              & 10.63               & 17.45                  & 77.03                  & 6.75               \\
Qwen2.5-7B-Instruct      & 27.93                & 37.00                 & 34.72              & 10.18               & 0.00                  & 76.34                  & 9.34               \\
Gemma-2-9B-IT         & 28.86                & 31.95                 & 42.14              & 14.77               & 0.23                  & 74.36                  & 9.74               \\
\hline
Stage-1-Llama         & 24.51                & 25.87                 & 26.32              & 7.83               & 19.26                  & 62.89                  & 4.88               \\
Stage-2-Llama         & 27.75                & 28.10                 & 24.64              & 7.72               & 19.56                  & 78.78                  & 7.74               \\
Merge-1-Llama         & 27.49                & 27.47                 & 26.22              & 8.28               & 19.79                  & 76.16                  & 7.04               \\
Merge-2-Llama         & 29.96                & 29.92                 & 28.78              & 9.96               & 19.94                  & 82.61                  & 8.54               \\
Aligned-SimPO-Llama      & 30.58                & 30.84                 & 34.31              & 8.39               & 26.59                  & 75.76                  & 7.61               \\
\hline
Llama-SEA-LION-8B-IT & 30.39                & 31.01                 & 29.47              & 10.40               & 22.58                  & 80.35                  & 8.54               \\
\hline
Stage-1-Gemma         & 29.88                & 33.34                 & 38.51              & 10.74               & 24.17                  & 56.87                  & \textbf{15.66}          \\
Stage-2-Gemma         & 33.48                & 34.67                 & 36.06              & 11.74               & 20.77                  & \textbf{83.00}              & 14.61               \\
Merge-1-Gemma         & 35.15                & 36.22                 & 41.42              & \textbf{15.32}          & 26.28                  & 82.09                  & 9.59               \\
Aligned-SimPO-Gemma      & 35.31                & \textbf{37.65}            & 42.38              & 14.99               & \textbf{27.79}             & 80.23                  & 8.82               \\
\hline
Gemma-SEA-LION-9B-IT & \textbf{35.43}            & 36.94                 & \textbf{43.39}          & 15.10               & 24.24                  & 81.85                  & 11.07              \\
\hline
\end{tabular}
}
\caption{Open LLM Leaderboard benchmarks across different instruct models of similar sizes.}
\label{tab:lm_eval-table}

\end{table*}

To understand the robustness and generalization of our proposed models, we conduct three studies as follows.
Section~\ref{subsec:cpt} evaluates the robustness of continual pre-training models using SEA-HELM and the Open LLM leaderboard.
In Section~\ref{subsec:sft}, we compare our instruction fine-tuning models with competitors in three benchmarks to demonstrate the generalization of our models.
Lastly, we discuss the design choice of our models in Section~\ref{subsec:ablation}.

\subsection{Continued Pre-Training Results} \label{subsec:cpt}

\noindent
\textbf{SEA performance}.
The CPT stage is primarily focused on gaining SEA language capabilities and knowledge. 
For the purpose of comparison against base and CPT models, as shown in Table~\ref{tab:cpt-sea-helm-table}, we observed a 6.05 and 7.19 average SEA-HELM performance increase over the \textit{Meta-Llama-3.1-8B} and \textit{Gemma-2-9B} for \textit{Llama-SEA-LION-8B} and \textit{Gemma-SEA-LION-9B}, respectively. 
We observed a much larger average increase with instruction following capabilities in particular, which we attribute to the fact that our CPT models are trained from the instruction models rather than from the base models. 
Moreover, in the average performance, we found that our Gemma-SEA-LION-9B models perform the best compared to other models.
This emphasizes a strong reason to perform CPT for improving the performance of SEA languages, rather than skipping the CPT and performing SFT directly.

\noindent
\textbf{English performance}.
For the English performance, as shown in Table~\ref{tab:base-lm-eval-table}, both CPT models also managed to perform competitively against the \textit{Meta-Llama-3.1-8B} and \textit{Gemma-2-9B base} models on the Open LLM Leaderboard benchmarks. 
This indicates that our choice of retraining with a proportion of 25\% English tokens has been beneficial in mitigating catastrophic forgetting, which has been shown to stem from CPT \citep{DBLP:conf/emnlp/ZhengPXQ0Z24}. 
Although our CPT models perform lower than Qwen and SeaLLMs on this benchmark, we outperform them on the SEA language instead, which is the main focus of this work.


%
%

\subsection{Instruction Fine-tuning Results}  \label{subsec:sft}

In this study, we compare our models with competitors on SEA-HELM, SEACrowd, and the Open LLM Leaderboard as follows.

\noindent
\textbf{SEA-HELM}. As shown in Table~\ref{tab:seahelm-posttrain-table}, the SEA-HELM benchmark performance demonstrates that our instruct models, \textit{Llama-SEA-LION-8B-IT} and \textit{Gemma-SEA-LION-9B-IT}, attain competitive performance in SEA languages, with \textit{Gemma-SEA-LION-9B-IT} achieving one of the highest average performances.
Moreover, we significantly improve the performance of Llama-3.1-8B-Instruct from 41.48 to 61.84 using \textit{Llama-SEA-LION-8B-IT}, while \textit{Gemma-SEA-LION-9B-IT} achieves 14.02 improvement points compared to Gemma-2-9B-IT.
Both \textit{Llama-SEA-LION-8B-IT} and \textit{Gemma-SEA-LION-9B-IT} outperform other SEA languages-focused LLMs, such as \textit{Sailor2-8B-Chat} and \textit{SEALLMs-v3-7B-Chat}, with an average score of 69.35 across all the languages covered by the SEA-HELM benchmark, apart from the SEA-MTBench tasks. 
This conforms with the previous results on the CPT models (Section~\ref{subsec:cpt}) that our CPT  model performs the best on SEA languages, resulting in the best performer in this experiment. 

\noindent
\textbf{SEACrowd}. Other than evaluating on some SEA languages like SEA-HELM, we also evaluated our model compared to competitors on 31 SEA indigenous languages using SEACrowd-NLU.
Note that, for this study, we use only the best settings of our models from the previous experiment (Table~\ref{tab:seahelm-posttrain-table}). 
As shown in Table~\ref{tab:lang-equity}, we observe a state-of-the-art result from \textit{Gemma-SEA-LION-9B-IT} by achieving 64.13 points on the NLU benchmark, while \textit{Llama-SEA-LION-8B-IT} improves its baseline from 49.94 to 55.10 points.
Moreover, the results from Figure~\ref{fig:sea-nlu-overall-results} also emphasize the robustness of our model by reaching more than 80 points on this benchmark, while SeaLLMs and Llama-3.1 have only a few cases where the performance exceeds 80 points.
These results emphasize the robustness of our models by achieving the state-of-the-art with a model parameter less than 10B on SEA benchmarks, including both traditional classical NLP benchmark (SEACrowd-NLU) and modern LLM benchmark (SEA-HELM).

\noindent
\textbf{English performance}. We also evaluate the performance of a widely used language, English, to observe a difference between the results of SEA and English.
The Open LLM Leaderboard performance is shown in Table~\ref{tab:lm_eval-table}.
Both \textit{Llama-SEA-LION-8B-IT} and \textit{Gemma-SEA-LION-9B-IT} performed competitively in English language, math, and reasoning tasks, with \textit{Gemma-SEA-LION-9B-IT} achieving the highest average score of 35.43. 
Moreover, we notice that the SEA models (Sailor and SeaLLMs) failed to perform on the English dataset.
This might be because these models are optimized for SEA languages during supervised fine-tuning, and English performance decreased as a result. 
In contrast, our models balance the performance between SEA and English knowledge, resulting in a high score for all benchmarks.

%

\subsection{Performance Analysis}  \label{subsec:ablation}
In this study, we discuss the performance improvement in each design decision of our models (Tables~\ref{tab:seahelm-posttrain-table} and ~\ref{tab:lm_eval-table}) as follows.

\noindent
\textbf{Stage 1: English instruction fine tuning} 
In Stage 1 IFT, the focus is predominantly on gaining general capabilities in math, code and general instruction following in the English language. Although our CPT models are based off of the instruct versions of \textit{Llama-3.1-8B}, the CPT process has eroded the instruction following capabilities (See Table~\ref{tab:lm_eval-table}). We observe an increase of 3.86 and 9.72 for \emph{Stage-1-Llama} and \emph{Stage-1-Gemma} respectively in English instruction following capabilities on the IFEval benchmark. We also observe an average increase of 7.9 for \emph{Stage-1-Llama} and 7.47 for \emph{Stage-1-Gemma} for the SEA-HELM benchmark. \\
\noindent
\textbf{Stage 2: Multilingual instruction fine tuning} 
In Stage 2 IFT, the focus is on multilingual and reasoning capabilities. By instruction fine tuning on SEA languages and higher complexity English instruction pairs, the Stage 2 models saw an average increase of 8.73 for \emph{Stage-2-Llama} and 10.1 for \emph{Stage-2-Gemma} over Stage 1 models on the SEA-HELM benchmark. \\
\noindent
\textbf{Merge 1: Combining Stage 1 and Stage 2}
Despite the significant gains observed in Stage 1 and 2, we observed that the effects of catastrophic forgetting from earlier stages could still be observed after Stage 2. In order to mitigate this, we merge Stage 1 and Stage 2 models into the CPT model, after which we we observed an average increase of 2.6 for \emph{Merge-1-Gemma}. We also observed an increase across all SEA-HELM benchmark tasks for \emph{Merge-1-Llama}. \\
\noindent
\textbf{Merge 2: Incorporating instruct models}
To reintroduce helpfulness, relevance and informativeness of responses observed in Llama 3.1 and Gemma 2 models, we perform further merges of open-source instruct models. While we observed significant increases in MT-Bench benchmark scores for Vietnamese and Thai, we also observed a slight degradation of average SEA-HELM performance as well as a slight degradation of Indonesian MTBench scores, which we view as acceptable tradeoffs for the significant performance increases in Vietnamese and Thai.\\
%
\noindent
\textbf{Alignment steps}
In the alignment step to align the models to human preference, we prioritize the SEA MTBench performance over the other SEA-HELM benchmark tasks. We observed a broad increase in SEA MTBench performances across all languages for both models. However, this comes with minor degradation of instruction following capabilities and overall Indonesian SEA-HELM performance. \\
The alignment step encourages longer, more helpful and sensitive responses but hurts performance on task-specific benchmarks and instruction following in some languages -- an issue we address in the next step.\\
%
\noindent
\textbf{Final merge: Combining aligned models}
To compensate for the capability degradation in the previous steps, we merge \emph{Merge-2-Llama} and \emph{Merge-1-Gemma} with \emph{Aligned-SimPO-Llama} and \emph{Aligned-SimPO-Gemma} and various open sourced pre-trained models describe in sections~\ref{sec:sft_llama_proc} and~\ref{sec:sft_gemma_proc} for their respective model families. 
For \textit{Llama-SEA-LION-8B-IT}, we observed a significant increase in average SEA-HELM performance (61.84) from the alignment stage (51.30), mainly from the increase in performance for the core tasks in SEA-HELM. This performance increase demonstrates the value of empirical selection of pre-trained models to be merged in based on each model's strengths and weaknesses to produce a far superior model.
For \textit{Gemma-SEA-LION-9B-IT}, it easily achieves higher performance compared to the \textit{Llama-SEA-LION-8B-IT} with fewer post training steps. We attribute this performance to the high performance of the base Gemma 2 model and also to the larger vocabulary size which have been demonstrated \citep{DBLP:journals/corr/abs-2406-16508} to produce better models.

\section{Related Works}
Recently, researchers have proposed large language models that support multilingual settings.
Llama~\cite{DBLP:journals/corr/abs-2407-21783} is the prior effort to release an open-source large language model for the research community to develop their own models.
Then, Qwen~\cite{DBLP:journals/corr/abs-2407-10671} and Gemma~\cite{DBLP:journals/corr/abs-2408-00118} introduced open-source LLMs that perform comparably or better than Llama with a larger amount of training data and many supported languages for these recent models.
Massively multilingual open-source models like Bloom \cite{DBLP:journals/corr/abs-2211-05100} and Aya \cite{DBLP:conf/acl/UstunAYKDOBSOKV24} also support a very wide range of languages, including some SEA languages.
Although these models demonstrate a robust performance in English benchmarks, they mostly underperformed on SEA benchmarks that tested for SEA languages, SEA knowledge and cultural understanding~\cite{DBLP:conf/emnlp/LoveniaMAMSAFMI24,susanto2025seahelmsoutheastasianholistic}, presumably due to a lack of language support for certain SEA languages or cultures.

In the SEA community, many works propose a large language model that is designed specifically for SEA languages by adding more SEA tokens in the training process, such as SeaLLMs~\cite{nguyen-etal-2024-seallms} and Sailor~\cite{sailor2report}.
However, the performance of these models is robust only on in-domain datasets or favors only some tasks (i.e., classical NLP datasets).
This is because the design choice in the pre-training or fine-tuning of these models is not well studied, e.g., performing a single SFT step with low-quality datasets written in some SEA languages, resulting in a slight improvement on SEA benchmarks.
To create a robust SEA LLM, we need to carefully balance language representation and design both pre-training and post-training (i.e., SFT, alignment, and model merging) for SEA contexts.

%
\section{Conclusion}
Despite the sizable population and language diversity in Southeast Asia, there remains a scarcity of resources and accurate linguistic and cultural representation with open-source LLMs. 
In this paper, we introduce \textit{Llama-SEA-LION-8B-IT} and \textit{Gemma-SEA-LION-9B-IT}, two multilingual LLMs comprehensively trained to achieve state-of-the-art performances in SEA languages, based on the Llama and Gemma family of LLMs.
SEA-LION represents the next advancement in the development of LLMs that explicitly supports SEA languages.
Both models are fully open-source and available for commercial use to increase accessibility and innovation in multilingual LLMs in Southeast Asia.
We will make our resources publicly available — including the dataset, training scripts, training checkpoints, and all fine-tuned models, even those that achieve state-of-the-art performance on the benchmarks — to establish solid baselines, ensure reproducibility, and support future research focused on culturally and professionally relevant SEA applications.

\section*{Acknowledgment}
This research is supported by the National Research Foundation, Singapore, under its National Large Language Models Funding Initiative. Any opinions, findings, and conclusions or recommendations expressed in this material are those of the author(s) and do not reflect the views of the National Research Foundation, Singapore.

\section*{Limitation}
Although we propose the state-of-the-art SEA LLMs, we found that the benchmark might not cover all the properties and languages we want to evaluate.
For example, SEA-HELM is a robustness benchmark, but only covers four languages.
SEACrowd is a benchmark that covers all SEA languages, but it is only classical NLP datasets (no chat or instruction following datasets).
We require a more holistic SEA benchmark that covers LLM-specific tasks written in all SEA languages.
However, with the current evaluation design choice, these benchmarks are the best design choice for current SEA research works.

Moreover, we conduct experiments using only 8 and 9 billion parameter models.
We argue that this is the most commonly used model size in real-world scenarios.
In addition, our method can and should also work with a higher or smaller model size since our proposed technique does not rely on the model size, as we demonstrated by applying the SFT and alignment techniques on both Llama and Gemma models.

%


%


\bibliography{reference_acl_final}

\newpage
\appendix
\onecolumn
\section{Appendix}
\label{sec:appendix}

\label{sec:datasets}

\subsection{Continued pre-training (CPT) data}

\noindent
\textbf{Existing data}: We utilize existing datasets as shown in Table~\ref{tab:cpt-data} (HuggingFace Datasets).

\noindent
\textbf{Other data}: As shown in Table~\ref{tab:cpt-data} (the other data section), the listed datasets contain data from a diverse range of domains, including news, books, articles, poems, etc. 

\begin{table*}[ht]
\captionsetup{justification=centering}
\centering
\scalebox{0.7}{
\begin{tabular}{lcc}
\hline
\multicolumn{3}{c}{\textbf{Continued Pre-training Data}}                                         \\ \hline
\textbf{Source (HuggingFace Datasets)} & \textbf{Languages} & \textbf{Size (Billions of Tokens)} \\ \hline
bigcode/the-stack-v2-dedup             & CODE               & 40                                 \\
allenai/dolma                          & EN                 & 37.5                               \\
HuggingFaceFW/fineweb-edu              & EN                 & 7.5                                \\
aisingapore/SEA-PILE-v1                & SEA                & 47.58                              \\
aisingapore/SEA-PILE-v2                & ID                 & 7                                  \\ \hline
\textbf{Source (Others)}               & \textbf{Languages} & \textbf{Size (Billions of Tokens)}          \\ \hline
VinBigData                             & VI                 & 16                                 \\
WangChanBERTa                          & TH                 & 8.5                                \\
Others - EN                            & EN                 & 5                                  \\
Others - SEA                           & SEA                & 30.92                              \\ \hline
\end{tabular}
}
\caption{List of datasets for the continued pre-training stage.}
\label{tab:cpt-data}
\end{table*}

\subsection{Stage 1 IFT data}

\begin{table*}[ht]
\centering
\captionsetup{justification=centering}
\scalebox{0.7}{
\begin{tabular}{lcc}
\multicolumn{3}{c}{\textbf{Stage 1 IFT Datasets}}              \\ \hline
\textbf{Source (HuggingFace Datasets)}           & \textbf{Languages} & \textbf{Size} \\ \hline
BAAI/Infinity-Instruct    & EN                 & 7,449,106     \\
nvidia/OpenMathInstruct-2 & EN                 & 2,000,000    \\ \hline
\end{tabular}
}
\caption{List of datasets for Stage-1-IFT. For BAAI/Infinity-Instruct dataset, any conversation that originally ended with a user turn has had that last turn removed.}
\label{tab:stage-1-ift-data}
\end{table*}

\subsection{Stage 2 IFT data}
\label{sec:stage-2-data}
\noindent
\textbf{Existing data}: We utilize existing datasets as shown in Table~\ref{tab:stage-2-ift-data} (HuggingFace Datasets).

\noindent
\textbf{Synthetic data}: As shown in Table~\ref{tab:stage-2-ift-data} (the generated part), we describe how to formulate synthetic data as follows

\begin{compactitem}
    \item qwen\_gemma\_synthetic datasets are generated first in English with Qwen 32B, utilizing an approach similar to Magpie. Instructions are then translated into the target language with Gemma 2 27B.
    \item llama\_gemma\_synthetic datasets are generated first in English with Llama 3.1 70B, utilizing an approach similar to Magpie \cite{DBLP:journals/corr/abs-2406-08464}. Instructions are then translated into the target language with Gemma 2 27B.
    \item gemma\_synthetic datasets are generated directly with Gemma 2 27B using Magpie \cite{DBLP:journals/corr/abs-2406-08464}.
    \item sea\_multilingual\_systemchat is a synthetic dataset translated with Gemma 2 27B from the English systemchat dataset.
    \item rewritten\_oasst is a dataset rewritten with Gemma 2 27B based on the English OASST dataset.
    \item rewritten\_helpsteer is a dataset rewritten with Gemma 2 27B based on the English Helpsteer dataset.
    
\end{compactitem}

\subsection{Helpfulness and preference alignment data}
\label{sec:helpfulness-and-preference-alignment-data}

As shown in Table~\ref{tab:sea-preference-data}, we use the princeton-nlp/gemma2-ultrafeedback-armorm as the source of the alignment data.
We then further re-scored with the reward model, nvidia/Llama-3.1-Nemotron-70B-Reward to create the SEA version. 
In particular, generated-gemma2-27b-seapref-nemotron-70b takes prompts from seald, wangchan\_thaiinstruct, and additional hand-written Southeast Asian cultural prompts collected from native speakers and then generates responses (with a varying temperature) from them with Gemma 2 27B. The responses are then scored with nvidia/Llama-3.1-Nemotron-70B-Reward, with the top-scoring response selected as chosen and vice versa, similar to princeton-nlp/gemma2-ultrafeedback-armorm.

\begin{table*}[h!]
\centering
\scalebox{0.7}{
\begin{tabular}{lcc}
\multicolumn{3}{c}{\textbf{Preference Data}}                                   \\ \hline
\textbf{Source (HuggingFace Datasets)}    & \textbf{Languages} & \textbf{Size} \\ \hline
princeton-nlp/gemma2-ultrafeedback-armorm & EN                 & 61,510        \\ \hline
\textbf{Source (Generated)}               & \textbf{Languages} & \textbf{Size} \\ \hline
generated-gemma2-27b-seapref-nemotron-70b & SEA                & 5,511         \\ \hline
\end{tabular}}
\caption{List of preference datasets used for the alignment stage.} 
\label{tab:sea-preference-data}
\end{table*}

\begin{table*}[ht]
\centering
\captionsetup{justification=centering}
\scalebox{0.7}{
\begin{tabular}{lcc}
\multicolumn{3}{c}{\textbf{Stage 2 IFT Datasets}}                                                                           \\ \hline
\textbf{Source (HuggingFace Datasets)}                                                 & \textbf{Languages} & \textbf{Size} \\ \hline
BAAI/Infinity-Instruct\textasciicircum{}*                                              & EN                 & 1,456,927     \\
HuggingFaceTB/smoltalk                                                                 & EN                 & 409,537       \\
allenai/tulu-3-sft-personas-math                                                       & EN                 & 149,960       \\
parinzee/seed-free-synthetic-instruct-thai-v1                                          & TH                 & 118,898       \\
HuggingFaceTB/smoltalk                                                                 & EN                 & 96,356        \\
HuggingFaceTB/smoltalk                                                                 & EN                 & 83,144        \\
arcee-ai/EvolKit-75K                                                                   & EN                 & 74,174        \\
AI-MO/NuminaMath-TIR                                                                   & EN                 & 72,441        \\
Post-training-Data-Flywheel/AutoIF-instruct-61k                                        & EN                 & 61,492        \\
argilla/ifeval-like-data                                                               & EN                 & 56,339        \\
HuggingFaceTB/smoltalk                                                                 & EN                 & 53,342        \\
ai2-adapt-dev/tulu\_v3.9\_wildjailbreak\_decontaminated\_50k                           & EN                 & 50,000        \\
ai2-adapt-dev/tulu\_v3.9\_synthetic\_finalresp\_wildguardmixtrain\_decontaminated\_50k & EN                 & 50,000        \\
allenai/tulu-3-sft-personas-math-grade                                                 & EN                 & 49,980        \\
allenai/tulu-3-sft-personas-code                                                       & EN                 & 34,999        \\
HuggingFaceTB/smoltalk                                                                 & EN                 & 34,424        \\
allenai/tulu-3-sft-personas-instruction-following                                      & EN                 & 29,980        \\
airesearch/WangchanThaiInstruct                                                        & TH                 & 25,014        \\
allenai/tulu-3-sft-personas-algebra                                                    & EN                 & 20,000        \\
arcee-ai/EvolKit-20k-vi                                                                & VI                 & 15,378        \\
allenai/coconot                                                                        & EN                 & 10,983        \\
ai2-adapt-dev/tulu\_v3.9\_sciriff\_10k                                                 & EN                 & 10,000        \\ \hline
\textbf{Source (Generated)}                                                            & \textbf{Languages} & \textbf{Size} \\ \hline
qwen\_gemma\_synthetic\_tamil                                                          & TA                 & 480,000       \\
qwen\_gemma\_synthetic\_thai                                                           & TH                 & 480,000       \\
qwen\_gemma\_synthetic\_indonesian                                                     & ID                 & 465,019       \\
qwen\_gemma\_synthetic\_vietnamese                                                     & VI                 & 465,019       \\
gemma\_synthetic\_indonesian                                                           & ID                 & 458,149       \\
gemma\_synthetic\_filipino                                                             & TL                 & 455,093       \\
gemma\_synthetic\_viet                                                                 & VI                 & 291,576       \\
gemma\_synthetic\_tamil                                                                & TA                 & 276,314       \\
gemma\_synthetic\_thai                                                                 & TH                 & 186,339       \\
gemma\_synthetic\_javanese                                                             & JV                 & 110,000       \\
gemma\_synthetic\_sudanese                                                             & SU                 & 110,000       \\
llama\_gemma\_synthetic\_thai                                                          & TH                 & 88,920        \\
llama\_gemma\_synthetic\_tamil                                                         & TA                 & 88,920        \\
llama\_gemma\_synthetic\_vietnamese                                                    & VI                 & 88,920        \\
llama\_gemma\_synthetic\_javanese                                                      & JV                 & 88,920        \\
llama\_gemma\_synthetic\_indonesian                                                    & ID                 & 88,920        \\
llama\_gemma\_synthetic\_filipino                                                      & TL                 & 80,000        \\
enrich\_27k                                                                            & SEA                & 27,463        \\
sea\_multilingual\_systemchat                                                          & SEA                & 1,903         \\
rewritten\_oasst                                                                       & SEA                & 841           \\
rewritten\_helpsteer                                                                   & SEA                & 838           \\ \hline
\end{tabular}
}
\caption{List of datasets for Stage-2-IFT.
}
\label{tab:stage-2-ift-data}
\end{table*}

\end{document}